\documentclass[conference]{IEEEtran}
\IEEEoverridecommandlockouts
\usepackage{cite}
\usepackage{amsmath,amssymb,amsfonts}
\usepackage{algorithmic}
\usepackage{graphicx}
\usepackage{textcomp}
\usepackage{xcolor}
\def\BibTeX{{\rm B\kern-.05em{\sc i\kern-.025em b}\kern-.08em
    T\kern-.1667em\lower.7ex\hbox{E}\kern-.125emX}}
\usepackage[table]{xcolor} % for \rowcolors
\usepackage{capt-of}       % for \captionof{table}

\usepackage{tcolorbox}
\usepackage[inline]{enumitem}
\usepackage{url}

\definecolor{promptbg}{RGB}{240, 248, 255}      % Light blue background
\definecolor{promptborder}{RGB}{70, 130, 180}   % Steel blue border

\newcounter{promptcounter}

\newenvironment{promptbox}[2]{%
    \refstepcounter{promptcounter}%
    \label{#1}%
    \begin{tcolorbox}[
        colback=promptbg,
        colframe=promptborder,
        coltitle=white,
        colbacktitle=promptborder,
        title={\textbf{Prompt \thepromptcounter: #2}},
        fonttitle=\sffamily\bfseries,
        rounded corners,
        left=4pt,
        right=4pt,
        top=6pt,
        bottom=6pt
    ]%
}{%
    \end{tcolorbox}%
}

\usepackage{framed}
\begin{document}

\title{PaVeRL-SQL: Text\textendash to\textendash SQL via Partial\textendash Match Rewards and Verbal Reinforcement Learning

}

% \author{\IEEEauthorblockN{Heng Hao\IEEEauthorrefmark{1}\IEEEauthorrefmark{2}, Wenjun Hu\IEEEauthorrefmark{1}\IEEEauthorrefmark{2}, Oxana Verkholyak\IEEEauthorrefmark{1}, Davoud Ataee Tarzanagh\IEEEauthorrefmark{1}, Baruch Gutow\IEEEauthorrefmark{1}, \\ 
% Sima Didari\IEEEauthorrefmark{1}, 
% Masoud Faraki\IEEEauthorrefmark{1}, Hankyu Moon\IEEEauthorrefmark{1} and Seungjai Min\IEEEauthorrefmark{1}}
% \IEEEauthorblockA{\IEEEauthorrefmark{1}Samsung SDSA, Mountain View CA, USA}
% \IEEEauthorblockA{\IEEEauthorrefmark{2}Equal contribution.\\
% Emails:  h.heng, wenjun.hu1, oxana.v, d.tarzanagh, baruch.g, s.didari, m.faraki, hankyu.m, seungjai.min @samsung.com}
% }

\author{\IEEEauthorblockN{Heng Hao\IEEEauthorrefmark{1}\IEEEauthorrefmark{2}, 
Wenjun Hu\IEEEauthorrefmark{1}\IEEEauthorrefmark{2}, 
Oxana Verkholyak\IEEEauthorrefmark{1}, 
Davoud Ataee Tarzanagh\IEEEauthorrefmark{1}, 
Baruch Gutow\IEEEauthorrefmark{1}, \\ 
Sima Didari\IEEEauthorrefmark{1}, 
Masoud Faraki\IEEEauthorrefmark{1}, 
Hankyu Moon\IEEEauthorrefmark{1}, 
and Seungjai Min\IEEEauthorrefmark{1}}
\IEEEauthorblockA{\IEEEauthorrefmark{1}Samsung SDSA, Mountain View, CA, USA}
\IEEEauthorblockA{\IEEEauthorrefmark{2}Equal contribution}
\IEEEauthorblockA{Emails: \{h.heng, wenjun.hu1, oxana.v, d.tarzanagh, baruch.g, s.didari, m.faraki, hankyu.m, seungjai.min\}@samsung.com}
}

\maketitle

\begin{abstract}
Text\textendash to\textendash SQL models allow users to interact with a database more easily by generating executable SQL statements from natural\textendash language questions. Despite recent successes on simpler databases and questions, current Text\textendash to\textendash SQL methods still suffer from low execution accuracy on industry\textendash scale databases and complex questions involving domain\textendash specific business logic. We present \emph{PaVeRL\textendash SQL}, a framework that combines \emph{Partial\textendash Match Rewards} and \emph{Verbal Reinforcement Learning} to drive self\textendash improvement in reasoning language models (RLMs) for Text\textendash to\textendash SQL. To handle practical use cases, we adopt two pipelines: (1) a newly designed in\textendash context learning framework with group self\textendash evaluation (verbal\textendash RL), using capable open\textendash and closed\textendash source large language models (LLMs) as backbones; and (2) a chain\textendash of\textendash thought (CoT) RL pipeline with a small backbone model (OmniSQL\textendash 7B) trained with a specially designed reward function and two\textendash stage RL. These pipelines achieve state\textendash of\textendash the\textendash art (SOTA) results on popular Text\textendash to\textendash SQL benchmarks—Spider, Spider~2.0, and BIRD. For the industrial\textendash level Spider2.0\textendash SQLite benchmark, the verbal\textendash RL pipeline achieves an execution accuracy 7.4\% higher than SOTA, and the CoT pipeline is 1.4\% higher. RL training with mixed SQL dialects yields strong, threefold gains, particularly for dialects with limited training data. Overall, \emph{PaVeRL\textendash SQL} delivers reliable, SOTA Text\textendash to\textendash SQL under realistic industrial constraints.  The code is available at {\color{blue}\url{https://github.com/PaVeRL-SQL/PaVeRL-SQL}}. 
\end{abstract}


\section{Introduction}
Structured Query Language (SQL) is the de facto interface to relational data. Text-to-SQL models translate natural-language questions into executable SQL, enabling non-experts to query complex databases. As large, heterogeneous schemas proliferate across industries—and as SQL dialects vary—the potential impact of reliable Text-to-SQL systems continues to grow. Recent systems advance this goal via structural bias and decoding constraints (e.g., RAT\textendash SQL, PICARD) or prompt/workflow design (e.g., DIN\textendash SQL, CHESS)~\cite{wang2019rat,scholak2021picard,pourreza2024din,talaei2024chess}.

Despite strong progress on academic benchmarks such as Spider~\cite{yu2018spider} and BIRD~\cite{li2024can}, execution accuracy (EX) remains fragile on industry-scale databases and harder queries that require multi-step reasoning and domain-specific business logic. The newer Spider~2.0~\cite{lei2024spider} further surfaces this gap by stressing realistic schema complexity and stricter evaluation.

Moving from benchmark settings to production environments reveals recurring challenges: (i) \textit{\underline{Complexity}}—target queries are long, compositional, and logic-heavy (multi-way joins, nested subqueries, window/aggregate functions), with Common Table Expressions (CTEs), aliases, and schema constraints compounding error propagation; (ii) \textit{\underline{Uncertainty}}—user questions can be ambiguous or under-specified, while schema/context at inference may be partial, noisy, or stale, requiring robust intent inference and schema grounding (see, for example, schema-linking and context-aware prompting)~\cite{kothyari-etal-2023-crush4sql,gao2023texttosqlempoweredlargelanguage}; (iii) \textit{\underline{Data sparsity}}—high-quality question–SQL pairs are scarce and uneven across domains and dialects, leaving long-tail operators and rare join patterns underrepresented; (iv) \textit{\underline{Generalization}}—supervised fine-tuning (SFT) and heuristic augmentation often overfit, degrading under distribution shift (new schemas, domains), and synthetic data is labor-intensive and brittle to prompt or schema changes; (v) \textit{\underline{Resource constraints}}—latency, cost, and governance concerns favor compact on-premises models over large general models, demanding efficient, auditable pipelines that meet service-level agreements (SLAs) and compliance requirements. Moreover, exact-match and execution-match metrics can provide \emph{sparse} or misleading signals in realistic settings~\cite{renggli2025fundamental}.

Recent advances in \emph{Reasoning} Language Models (RLMs) trained with reinforcement learning (RL)—for example, o1~\cite{openai2024openaio1card} and DeepSeek\textendash R1~\cite{guo2025deepseekr1}—show that verifiable feedback can substantially improve multi-step reasoning, aligning with chain-of-thought evidence~\cite{suzgun2022challengingbigbenchtaskschainofthought} and the heuristic that supervised fine-tuning memorizes while RL generalizes~\cite{chu2025sft}. Text-to-SQL is naturally amenable to such feedback via execution signals (see also early execution-based RL in Seq2SQL~\cite{zhong2017seq2sql}), yet rewards remain \emph{sparse} (binary success/failure) and dialect coverage uneven, complicating training and deployment. In this work we operationalize RL with group-relative policy optimization (GRPO)~\cite{shao2024grpo} while addressing reward sparsity through denser, partial-match signals.
\vspace{.25cm}

\noindent\textbf{Contributions.} We present \textbf{PaVeRL–SQL} (Partial–Match Rewards and Verbal RL for Text–to–SQL), an RL framework that makes the following contributions:
\begin{enumerate}[label=\textnormal{\Roman*.}, leftmargin=*, topsep=1pt, itemsep=1pt]
    \item We design, build, and evaluate two complementary pipelines tailored to deployment constraints: (i) a \emph{verbal self\textendash evaluation} in\textendash context workflow that samples executable candidates until it collects \(K=10\) valid SQLs per query and uses the backbone large language model to rank and select the final query; and (ii) a \emph{chain\textendash of\textendash thought} reinforcement\textendash learning pipeline that trains a small on\textendash premises backbone (OmniSQL\textendash 7B) end\textendash to\textendash end with execution feedback. Together, these pipelines achieve state\textendash of\textendash the\textendash art (SOTA) results on Spider, Spider2.0\textendash SQLite, and BIRD.
    
    \item We introduce evaluation metrics and reward shaping that capture \emph{fractional} correctness via column\textendash level fractional execution accuracy ($\text{EX}_f$) alongside binary execution accuracy ($\text{EX}_b$), providing denser, more informative signals than 0/1 exact match and improving training stability.
    \item We develop a cost\textendash effective two\textendash stage group\textendash relative policy optimization (GRPO) schedule—best\textendash checkpoint restart followed by cosine decay—that improves stability and sample/compute efficiency, reaching strong accuracy within a modest training budget (e.g., \(\leq 20\) epochs) without second\textendash order methods.
    
    \item We demonstrate mixed\textendash dialect training that transfers across SQL dialects and yields substantial gains for low\textendash resource dialects, strengthening cross\textendash dialect generalization under realistic industrial constraints.
\end{enumerate}

\section{Related Work}
 \paragraph{From Prompting \& Fine-Tuning to RL in Text-to-SQL} Text-to-SQL systems have advanced through (i) model adaptation and (ii) prompt engineering. Fine-tuning methods—RAT-SQL~\cite{wang2019rat}, PICARD~\cite{scholak2021picard}, RESDSQL~\cite{li2023resdsql}, CodeS~\cite{CodeS}, and PSM-SQL~\cite{yang2025psm}—inject schema structure and enforce decoding constraints. Prompt-driven approaches—ACT-SQL~\cite{zhang2023act}, DIN-SQL~\cite{pourreza2024din}, CHESS~\cite{talaei2024chess}, CHASE-SQL~\cite{pourreza2024chase}—use carefully designed instructions for zero/few-shot generalization. Other innovations include semantic schema linking~\cite{kothyari-etal-2023-crush4sql}, context-aware prompting~\cite{gao2023texttosqlempoweredlargelanguage}, modular query breakdown~\cite{pourreza2024din}, memory-based self-correction~\cite{shinn2024reflexion}, and confidence calibration~\cite{ramachandran2024texttosqlcalibrationneedask,tian2023justaskcalibrationstrategies}. These techniques deliver strong results on Spider~\cite{yu2018spider} and BIRD~\cite{li2024can}, but scaling to the complexity of Spider~2.0~\cite{lei2024spider} remains challenging—motivating RL for complicated industry-level databases and convoluted user questions involving domain-specific business logic.
 
 \paragraph{Reasoning LMs as Bases for RL} Reasoning Language Models (RLMs) trained with RL can produce deliberate chain-of-thought (CoT) traces, improving multi-step reasoning~\cite{suzgun2022challengingbigbenchtaskschainofthought, openai2024openaio1card}. Empirically, \emph{SFT memorizes, RL generalizes}~\cite{chu2025sft}, and large-scale RL boosts math/code performance~\cite{chen2025acereason}. RL is sensitive to base-model strength: weak cold starts hinder RL efficiency~\cite{guo2025deepseekr1}. Recent strong bases (o1~\cite{openai2024openaio1card}, DeepSeek R1~\cite{guo2025deepseekr1}, Qwen3~\cite{yang2025qwen3}, Phi-4~\cite{abdin2025phi4}) enable more reliable RL for Text-to-SQL. Two-stage training (e.g., DeepSeekLLM~\cite{deepseek2024deepseekllm}) can also prime models for reasoning; however, two-stage RL has not been thoroughly explored for Text-to-SQL. 
 
 \paragraph{RL for Text-to-SQL} A growing body of work applies RL to SQL generation. \textsc{Seq2SQL}~\cite{zhong2017seq2sql} uses policy gradients with execution feedback for clause refinement. \textsc{ICRL}~\cite{toteja2025context} introduces retrieval-based in-context learning with reward signals. \textsc{SQL-RL-GEN}~\cite{anonymous2024llmbased} and \textsc{STaR-SQL}~\cite{he2025star} explore evolutionary search and reward shaping. \textsc{DeepSQL\textendash R1}~\cite{kumar2025deepsql} targets efficiency via LoRA and quantization. \textsc{Reasoning\textendash RL}~\cite{pourreza2025reasoning} augments schema linking with similarity-based rewards. \textsc{SQL\textendash R1}~\cite{ma2025sqlr1} advances multi-table and nested queries using dynamic rewards, while \textsc{Arctic\textendash Text-to-SQL\textendash R1}~\cite{yao2025arcticsql} emphasizes execution correctness and syntax validity.

\paragraph{Reward Design} Execution accuracy provides a natural but \emph{sparse} reward in Text-to-SQL, which can impede policy optimization. Composite rewards have been proposed: LLM-as-a-Judge, syntax checks, schema-linking, and $n$-gram similarity~\cite{pourreza2025reasoning}; and format/execution/result/length components~\cite{ma2025sqlr1}. Recent simplifications focus on execution and syntax validity~\cite{yao2025arcticsql}. However, most approaches treat correctness as exact match against the ground truth, ignoring \emph{degrees} of partial correctness in results. Our work introduces a partial-match execution reward that supplies denser, more informative signals, stabilizing RL and improving learning efficiency. Moreover, we address multi-dialect training within a single RL pipeline, whereas prior RL methods are typically per-dialect, limiting cross-dialect generalization. 

\paragraph{Agents \& Workflows as RL Substrate} Agentic coding systems combine reasoning, tool use, and iterative correction~\cite{yao2023react, chen2023teaching, yang2024intercode}. ReAct~\cite{yao2023react} integrates reasoning with actions; Spider Agent~\cite{lei2024spider} adapts this to database tasks. Multi-agent/workflow strategies~\cite{chen2024coder, shinn2024reflexion, wang2023plan} assist complex planning, though domain-specific methods often outperform general agents~\cite{xia2024agentless}. ReFoRCE~\cite{deng2025reforce} scales to Spider~2.0 with iterative exploration and schema compression. In our setting, such agentic infrastructure is complementary: it provides an environment where RL-trained Text-to-SQL policies can execute, receive feedback, and improve, but our contribution is \emph{model-centric RL training}, not orchestration logic.

\section{Methodology}
We introduce \textbf{PaVeRL\textendash SQL} \emph{(Partial\textendash Match Rewards and Verbal RL for Text\textendash to\textendash SQL)}, a two\textendash track approach that balances deployability and accuracy: a \emph{verbal self\textendash evaluation} workflow that treats a backbone LLM as both generator and judge, and a \emph{chain\textendash of\textendash thought (CoT) reinforcement learning} pipeline trained with group\textendash relative policy optimization (GRPO)~\cite{shao2024grpo}. Both tracks leverage verifiable execution feedback in the spirit of recent reasoning systems~\cite{openai2024openaio1card,guo2025deepseekr1} and are evaluated with a denser, partial\textendash match criterion to complement exact/execution match metrics~\cite{renggli2025fundamental}.

We designed and evaluated two specific pipelines for different use cases of Text-to-SQL to optimize performance and cost depending on available resources.

\subsection{Evaluation}
\label{sec:eval}
We adopted the official evaluation methodologies for each dataset to calculate the accuracy of the SQL execution (EX) generated in all test cases \cite{wang2019rat,scholak2021picard,pourreza2024din,talaei2024chess,lei2024spider,yu2018spider}. However, we also observed that there are obvious limitations of these evaluations, because these metrics compare the exact match of the execution results between the golden SQL and the generated SQL. These limitations of existing evaluation metrics are also well documented in other works~\cite{renggli2025fundamental}. Moreover, in many practical scenarios, the result can be counted as correct even if it does not exactly match the ground truth, provided that all the requested information is reflected in the execution result table. This can happen due to the vagueness of the user questions, differences in database setting, etc.

Considering the nature of SQL statements, we developed a new evaluation metric, which compares the golden result table with the generated SQL execution result table column by column. We report two complementary measures computed by comparing the generated result table $\hat{T}$ against the golden table $T$ \emph{column\textendash by\textendash column}:
        \begin{enumerate}
            \item \textbf{Binary execution accuracy} $\text{EX}_b\!\in\!\{0,1\}$ assigns, for each sample, a label as either correct (1) or incorrect (0). The criterion for $\text{EX}_b$ is less strict than an exact match. If the execution result table contains all the information from the golden table and the shape difference (the counts of extra columns) is less than $\tau$ (we use $\tau\!=\!5$ as default, tunable per application), it is still considered correct. %Here, the number 5 is a hyperparameter that can be adjusted depending on the use cases. (HH comment: duplicate to previous sentence) 
            Lower $\tau$ values mean tighter evaluation and higher values mean looser evaluation. Formally, if $|\mathrm{Cols}(\hat{T})\setminus \mathrm{Cols}(T)|<\tau$, we still count the sample as correct.
            \item \textbf{Fractional execution accuracy} $\text{EX}_f\!\in\![0,1]$ measures, for each sample, the proportion of the result column matching the golden result, with continuous values from 0 to 1. It averages per\textendash column matches: each column in $T$ contributes $1$ if its values match in $\hat{T}$ (up to row ordering and formatting), and $0$ otherwise; extra columns neither help nor hurt.
        \end{enumerate}
Together, $\text{EX}_b$ and $\text{EX}_f$ supply denser signals during analysis and training while remaining faithful to execution semantics.

\subsection{Verbal RL}
\label{sec:vRLmethod}

\textbf{Use Case 1}: The database is very similar to general databases in the public data, or training in-house Text-to-SQL models is not feasible due to hardware restriction or cost.

We developed a verbal self-evaluating RL pipeline inspired by Group Relative Policy Optimization (GRPO)~\cite{shao2024grpo} training process, Reflexion~\cite{shinn2024reflexion} and ReAct~\cite{yao2023react}. For settings where training an in\textendash house model is infeasible or the target schema is close to public distributions, we use this verbal self\textendash evaluation pipeline that approximates GRPO's groupwise preference signal and self\textendash reflection/act paradigms.

First, for each user question, we sample SQL candidates with the backbone LLM using \emph{Prompt}~\ref{sqlgenprompt}. We execute each candidate to check validity and repeat sampling until we collect $K=10$ executable SQLs or reach a cap of 200 attempts. If, after 200 attempts, we still have fewer than $K$ executable candidates, we proceed with the candidates collected so far. We then ask the same LLM to score these candidates using the scoring prompt (Appendix~\ref{app:vRLprompt}). For each SQL candidate 20 scores will be generated, and the final score of the SQL will be the mean of the 20 scores. We select the highest score SQL, and if multiple candidates tie, we break ties uniformly at random. The selected highest-scoring SQL is returned as the final output. 

\begin{figure}[t]
\centerline{\includegraphics[width=0.9\columnwidth]{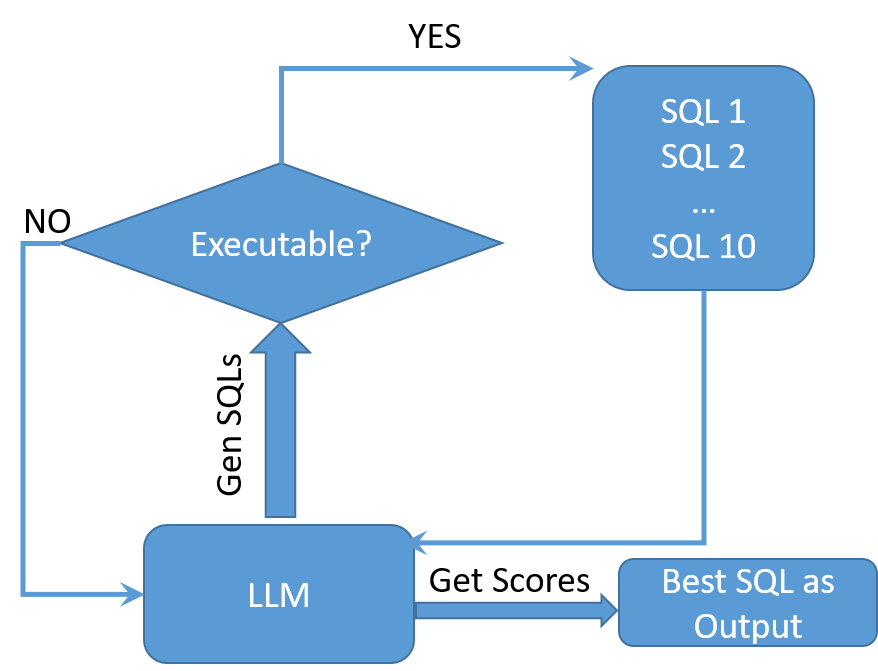}}
\caption{\textbf{Verbal RL Pipeline:} A generate-and-judge workflow that approximates GRPO without gradient updates. For each natural language question, the pipeline (1) samples SQL candidates from the backbone LLM using the generation prompt until collecting $K=10$ executable SQLs or reaching 200 attempts, (2) executes each candidate to verify validity, (3) uses the same LLM to score all executable candidates via the scoring prompt, and (4) selects the highest-scoring SQL as the final output.}
\label{fig:vRL}
\end{figure}

This "generate\textendash and\textendash judge" loop approximates a group\textendash relative signal without gradient updates and consistently improves over zero\textendash shot prompting baselines~\cite{talaei2024chess,lei2024spider2}. The entire pipeline is shown in Figure~\ref{fig:vRL}.

\subsection{CoT RL}
\label{sec:cotrl}
\textbf{Use Case 2}: There is a requirement to use small on-prem Text-to-SQL models and compute resource is available to train them. If access to the database is limited either due to the data security, or the cost of accessing is very high due to speed, size or any other factor of the database, it is recommended to train a one-shot SQL generator model based on CoT RL training with verifiable output rewards model (ORM).

When on\textendash prem inference and stronger robustness are required, we train a small backbone with GRPO using verifiable execution rewards.

\subsubsection{RL Approach}
For the CoT RL training, we adopt GRPO~\cite{shao2024grpo}, which has been shown to be very successful in many verifiable rewards tasks (e.g. math and code) and therefore is ideal to be extended to Text-to-SQL task. For an optimum training effect, a supervised finetuning (SFT) warm up is usually suggested~\cite{guo2025deepseekr1}. OmniSQL models~\cite{li2025omnisql} are SFT trained on Qwen2.5-coder models~\cite{hui2024qwen25coder} using SynSQL-2.5M~\cite{li2025omnisql} data, which is an ideal choice as the starting base model for GRPO (after the SFT warmup stage). We initialize from OmniSQL\textendash 7B~\cite{li2025omnisql}, an SFT\textendash primed coder model suited for SQL generation, and condition on \emph{Prompt}~\ref{sqlgenprompt} that includes dialect, schema, optional context, and question.

\paragraph{RL objective}
GRPO training proceeds as follows: using the prompt $P$ that contains the question, the database schema, and the relative context as input, the policy model $\pi_\theta$ (parameterized by $\theta$) will generate a set $G$ of SQL candidates (aka rollouts), $\{o_{1}, ..., o_{G}\}$. Each generated SQL is executed and the execution result is compared to the golden SQL execution result using the defined reward function (described in detail in Section~\ref{sec:rm}). 

After executing each $o_i$ and scoring it, GRPO updates the policy using group\textendash relative advantages $A_i$. By checking the relative performance of the SQL candidates within the group, the policy is updated in accordance with the following objectives:
\begin{equation*}  
\begin{split}
\mathcal{J}_{\text{GRPO}}(\theta) & := \mathbb{E}\bigg[  
  \frac{1}{G} \sum_{i=1}^G \min\left(r_i A_i, \; \mathrm{clip}(r_i, 1-\epsilon, 1+\epsilon) A_i\right)  
\bigg]  \\
&- \beta \mathrm{D_{KL}}(\pi_\theta \| \pi_{ref}),
\end{split}
\end{equation*} 
where $P$ is the input prompt (dialect, schema, optional context, and question); $\pi_\theta$ is the current policy parameterized by $\theta$; $G$ is the number of rollouts per prompt; $o_i$ is the $i$th generated SQL; $r_i = \frac{\pi_\theta(o_i|P)}{\pi_{\theta_{\text{old}}}(o_i|P)}$ is the likelihood ratio of generating output $o_i$ under the new policy $\pi_\theta$ compared to the old policy $\pi_{\theta_{\text{old}}}$, $A_i$ is the group relative advantage of each output, $\epsilon>0$ is the PPO\textendash style clipping parameter, $\beta\ge 0$ weights the KL regularizer, and $\mathrm{D_{KL}}$ is a KL-divergence penalty to keep the policy close to the reference policy $\pi_{\text{ref}}$ (the SFT checkpoint). The expectation is over training prompts/minibatches. This mirrors successful verifiable\textendash reward training in math/code~\cite{openai2024openaio1card,guo2025deepseekr1} and prior execution\textendash based RL for SQL~\cite{zhong2017seq2sql}.

\paragraph{Two\textendash stage schedule}
Inspired by DeepSeekLLM~\cite{deepseek2024deepseekllm}, we adopt the two-stage training strategy to achieve optimal RL training from small number of training epochs (aka training budget). However, we made a change to the strategy as follows:   \textit{\underline{\textbf{Stage one}}}: For the first 10 epochs, apply the most commonly used learning rate schedule, which starts with initial 3\% steps of flat low learning rate to warm-up, followed by a linear increase to the pre-defined max learning rate for the 3\% to 10\% steps, then ending with a slow cosine decay of the learning rate for the rest of the steps. \textit{\underline{\textbf{Stage two}}}: Always start from the best model (highest greedy decoding accuracy) of stage one. The best model of stage one can appear at any steps (beginning, middle or end). Unlike typical two-stage training, the starting point of stage two is not constrained to the final model of stage one. We also need to observe the accuracy curve of stage one models:
\begin{enumerate}
\item If the accuracy curve increases then starts to show signs of plateauing, raise the stage two starting learning rate to the pre-defined max, followed by a slow cosine decay.
        \item If the accuracy curve fluctuates without showing a clear trend, reduce the starting learning rate of stage two by half (of the best model learning rate), followed by a slow cosine decay. 
    \end{enumerate}

Using the above training strategy, we gain satisfactory accuracy level within 20 epochs of training, leading to significant reduction of the training time and cost.

\subsubsection{Reward Function}
\label{sec:rm}
We defined a reward function that encourages the promotion of $\text{EX}_f$, that is, the fraction of correct columns in the result table. To reduce sparsity while preserving verifiability, we reward \emph{execution} and scale payoff by partial correctness. To save training time, we execute each golden SQL to get the golden result table before the GRPO training starts. Then during the training we only compare the rollout SQL execution result table to the golden result table. 

Let $\text{EX}_f\!\in\![0,1]$ denote the fractional column\textendash match score from Section~\ref{sec:eval}. The rewards are defined as follows: 

\begin{equation*}  
R =   
\begin{cases}  
10 \cdot \text{EX}_f,   & \text{if SQL executes and returns results;} \\  
0.5, & \text{if SQL executes but is incorrect;} \\  
0,   & \text{otherwise (if SQL fails to execute).}  
\end{cases}  
\end{equation*}  

During our experiments, we tested various reward function designs and observed that applying both reward (positive value) and punishment (negative value) causes confusion during GRPO training, leading to unstable results. Empirically, non\textendash negative rewards stabilize GRPO updates; mixing large positive and negative signals degraded convergence. The above reward function focuses on execution correctness and syntax validity. Moreover, it gives a continuous reward score based on the correctness fraction. This reward function contributed to a significant performance gain in our GRPO RL training and aligns with recent Text\textendash to\textendash SQL RL that centers execution and syntax validity~\cite{yao2025arcticsql,ma2025sqlr1} while adding a graded notion of correctness that improves sample efficiency.

\section{Experiments and Results}
We present empirical results from two practical Text-to-SQL pipelines under different deployment scenarios and compare them with SOTA approaches on the popular Spider~\cite{yu2018spider}, Spider~2.0~\cite{lei2024spider2}, and BIRD~\cite{li2024bird} benchmarks. We report three metrics: (1) the official execution accuracy (EX) defined by each dataset; (2) binary execution accuracy ($\text{EX}_b$); and (3) fractional execution accuracy ($\text{EX}_f$). All three metrics are described in Section~\ref{sec:eval}.

\subsection{Datasets and Pre-processing}
For ease of comparison, we focus on the SQLite dialect first and present results for another dialect in Section~\ref{sec:mix}.

\subsubsection{Spider, Spider~2.0\textendash SQLite, and BIRD Datasets} %\da{Spider2SQLite or Spider~2.0\textendash SQLit }

The Spider~\cite{yu2018spider} dataset consists of 8{,}659 training question–SQL pairs from 146 databases. We filtered out questions with missing columns and incorrect SQL (typically misuse of table abbreviations), resulting in 8{,}648 pairs. We then split these into 8{,}000 for training and 648 for validation for RL training. The trained models are evaluated on the development set (1{,}034 question–SQL pairs from 20 databases).

\begin{table}[ht]
\centering
\caption{Training and evaluation dataset statistics for PaVeRL-SQL experiments.}
\label{tbl:datatbl}
\rowcolors{2}{gray!15}{white}
\resizebox{7.5cm}{!}{%
\begin{tabular}{lcc}
\hline
\textbf{Training} & \textbf{\# Que} & \textbf{\# DB} \\ \hline
Spider\_train   & 8000  & 142 \\
Spider\_val     & 648   & 5 \\
BIRD\_train     & 8600  & 67 \\
BIRD\_val       & 429   & 2 \\
SynSQL\_train   & 10000 & 7387 \\
SynSQL\_val     & 175   & 109 \\
\hline
\rowcolor{white}\textbf{Testing} & \textbf{\# Que} & \textbf{\# DB} \\ \hline
Spider\_dev             & 1034 & 20 \\
Spider2.0\textendash SQLite & 135  & 30 \\
BIRD\_dev               & 1534 & 11 \\
BIRD\_dev\_sub$^{\mathrm{a}}$ & 147  & 11 \\
\hline
\multicolumn{3}{l}{$^{\mathrm{a}}$Small subsample for cost efficient inference.}
\end{tabular}
}

\end{table}

The BIRD~\cite{li2024bird} dataset contains 9{,}428 training question–SQL pairs from 69 databases. After removing samples with missing columns/tables or incorrect SQL, we obtained 9{,}029 pairs from 68 databases, split into 8{,}600 for training and 429 for validation. The BIRD development set includes 1{,}534 question–SQL pairs from 11 databases. Following CHESS~\cite{talaei2024chess}, we select a representative subsample (147 question-SQL pairs from 11 databases)\footnote{https://github.com/ShayanTalaei/CHESS/blob/main/data/dev/

 sub\_sampled\_bird\_dev\_set.json} for cost efficiency.
% we also use a representative subsample (147 question–SQL pairs from 11 databases)\footnote{https://github.com/ShayanTalaei/CHESS/blob/main/data/dev/sub\_sampled\_bird\_dev\_set.json} for cost efficiency.
Spider~2.0~\cite{lei2024spider2} contains 135 questions from 30 databases in SQLite dialect; we use this set only for evaluation.

To obtain a general-purpose Text-to-SQL model not limited by domain or difficulty, we added a third training set from SynSQL-2.5M~\cite{li2025omnisql}. We randomly selected 10K question–SQL pairs with balanced difficulty, covering 7{,}387 databases. We excluded \textit{multi-turn dialogue} style questions, as single-turn CoT RL is not ideal for this setting, and defer multi-turn modeling to future work. We also selected 175 question–SQL pairs from 109 databases as a validation set. For this dataset, we ignore the provided gold CoT and allow RL to fully explore solutions. The details are summarized in Table~\ref{tbl:datatbl}.

\subsubsection{Database Schema Details and Context Information}
During our experiments, we found that augmenting the database schema with lightweight data profiles—such as primary/foreign keys, min/max values for numeric columns, and the top-3 most frequent values for text columns—substantially improves SQL generation, especially for filtering and joins. Accordingly, we curate and use enriched metadata that includes these signals in addition to the canonical schema. An example schema string is shown in Appendix~\ref{app:dbschema}.

For BIRD, OmniSQL~\cite{li2025omnisql} segments long-form context into per-column descriptions and inlines them into the schema string. We adopt the same approach and observe meaningful gains. Because some BIRD databases are highly complex, the full schema will not fit in the model’s context window during RL training. Following OmniSQL~\cite{li2025omnisql}, we therefore include (i) all columns referenced by the gold SQL, (ii) all primary/foreign key columns, and (iii) a small set of randomly sampled additional columns. For the SPIDER and SynSQL10K training sample, we directly included all the database schema strings described above. For some SynSQL10K samples that have context information, we just included them as separate paragraphs in the prompt.  For testing, we always use the full database schema string. The prompt for generating the SQL is shown in Prompt~\ref{sqlgenprompt}.

\begin{promptbox}{sqlgenprompt}{SQL Generation Prompt}
Task Overview:

You are a powerful text-to-SQL model. Below, you are provided with a database schema and a natural language question. Your task is to understand the schema and generate a valid SQL query to answer the question.
\\\\
Database Engine:\\
\{dialect\}
\\
Database Schema:\\
\{dbschema\}
\\
Context:\\
\{context\}
\\
Question:\\
\{question\}
\\\\
Instructions:\\
- Please use the minimum number of tokens required to provide a SQL statement and use sql functions for \{dialect\} database.\\
- Make sure you only output the information that is asked in the question. If the question asks for a specific column, make sure to only include that column in the SELECT clause, nothing more.\\
- The generated query should return all of the information asked in the question without any missing or extra information.\\
- Please think through the steps of how to write the query with minimum number of tokens.\\
\\
Output Format:\\
In your answer, please enclose the thinking block with less than 1600 tokens followed by a code block with the generated SQL code:
\begin{verbatim}
<think>
-- Your brief thinking
</think>
```sql
-- Your SQL query
```
\end{verbatim}

Please DO NOT generate any explanation to the final SQL code solution.
\end{promptbox}

%%%%%%%%%%%%%%%%%%%%%%%%%%%%%%%%%%%%%%%%%%%%%%%%%%%
\subsection{Verbal RL Pipeline}
As described in Section~\ref{sec:vRLmethod} and shown in Figure~\ref{fig:vRL}, we evaluated the verbal RL pipeline using several recent open- and closed-source LLMs as backbones: GPT-5 mini\footnote{https://platform.openai.com/docs/models/gpt-5-mini} (`gpt-5-mini-2025-08-07'), %Claude Opus~4.1\footnote{https://docs.anthropic.com/en/docs/about-claude/models/overview} (`claude-opus-4-1-20250805'), 
gpt-oss-20B and gpt-oss-120B~\cite{openai2025gptoss120bgptoss20bmodel}, Qwen3-30B-A3B-Instruct-2507~\cite{yang2025qwen3}, and Qwen3-Coder-30B-A3B-Instruct~\cite{yang2025qwen3}. 
We tested on the Spider~2.0 SQLite subset (135 samples) and the BIRD dev subsample (147 samples). For time and cost considerations, we did not evaluate on the full BIRD dev set. For comparison, we also report zero-shot inference and SOTA pipelines with code that we can reproduce (CHESS~\cite{talaei2024chess} and SpiderAgent~\cite{lei2024spider2}) using official execution accuracy (EX), binary execution accuracy ($\text{EX}_b$), and fractional execution accuracy ($\text{EX}_f$) from Section~\ref{sec:eval}.

On Spider2.0–SQLite, our simple verbal RL pipeline reaches 37\% EX—\textbf{+12.6} points over SpiderAgent and \textbf{+7.4} over CHESS with the same backbone (GPT-5 mini). In general, the pipeline substantially outperforms vanilla zero-shot prompting across backbones. Prompts are provided in Appendix~\ref{app:vRLprompt}; detailed results are in Table~\ref{tbl:verbalRL}.

% \begin{figure*}[h]
% \centerline{\includegraphics[width=\textwidth]{output.png}}
% \caption{CoT RL Training results of Greedy EX, learning rate, and mean rewards from left to right. Blue and green dots are from stage 1 and 2, respectively. The datasets are Spider\_dev, Spider2SQLite, and BIRD\_dev from top to bottom.}
% \label{fig:cotrl}
% \end{figure*}

% ===== Table 1: Verbal RL (keeps label tbl:verbalRL) =====
\begin{table}[t]
\centering
\setlength{\tabcolsep}{3pt}
\caption{\label{tbl:verbalRL}\textbf{Verbal RL Execution Accuracy (\%) Compared to 0 Shot and Other Pipeline}}
\rowcolors{2}{gray!15}{white}
\resizebox{\columnwidth}{!}{%
\begin{tabular}{l|l|ccc|ccc}
\hline
&&\multicolumn{3}{c|}{Spider2.0--SQLite} & \multicolumn{3}{c}{BIRD} \\
\hline
Method & Backbone LLM & EX & $\text{EX}_b$ & $\text{EX}_f$  & EX & $\text{EX}_b$ & $\text{EX}_f$ \\
\hline
Agent$^{a}$~\cite{lei2024spider2} &  GPT-5 mini & 24.4 & 23.0 & 35.4 & 29.9 & 33.3 & 42.5 \\
CHESS$^{b}$~\cite{talaei2024chess} &  GPT-5 mini & 29.6 & 29.6 & 42.6 & 61.2 & 57.8 & 65.0\\
\hline
PaVeRL & GPT-5 mini  & \textbf{37.0} & \textbf{34.1} & \textbf{49.9} &  55.8 & 59.9 & 64.5 \\
& GPT OSS 20B & 28.1 & 27.4 & 39.5 & 53.1 & 60.5 & 64.9\\
& GPT OSS 120B & 33.3 & 32.6 & 48.5 & 54.4& 60.5& 64.8\\
& Qwen3-30B$^c$ & 11.1 & 11.1 & 21.2 & 53.7 & 53.7 & 57.3\\
& Qwen3-Coder-30B$^d$ & 20.0 & 18.5 & 30.2 & 60.5 & 64.6 & 67.3 \\
\hline
0 shot$^e$& GPT-5 mini & 33.1 & 28.1 & 42.1 & 55.1 & 55.8 & 60.5 \\
& Claude Opus 4.1 & 31.9 & 30.4 & 42.7 &  59.2 & 64.6& 68.1\\
& GPT OSS 20B & 8.89 & 8.89 & 10.8 & 51.0 & 54.4 & 59.0\\
& GPT OSS 120B & 18.5 & 14.8  & 21.7 & 54.4 & 57.8 & 61.6\\
& Qwen3-30B$^c$ & 11.1 & 10.4 & 16.3 & 46.9 & 56.5 & 59.1\\
& Qwen3-Coder-30B$^d$& 2.2 & 2.2 & 6.2 & 15.0 & 15.6 & 16.3\\
\hline
\end{tabular}}
\par\vspace{2pt}\footnotesize
\begin{minipage}{\columnwidth}
\raggedright
a. Agent = SpiderAgent~\cite{lei2024spider2}. We strictly follow the authors' official GitHub.\\
b. We strictly follow the CHESS~\cite{talaei2024chess} official GitHub.\\
c. Qwen3-30B = Qwen3-30B-A3B-Instruct-2507.\\
d. Qwen3-Coder-30B = Qwen3-Coder-30B-A3B-Instruct.\\
e. We use Prompt~\ref{sqlgenprompt} with enriched database schema as in Appendix~\ref{app:dbschema}.
\end{minipage}
\end{table}

% ===== Table 2: CoT-RL (keeps label tbl:cotRL) =====
\begin{table}[t]
\centering
\setlength{\tabcolsep}{3pt}
\caption{\label{tbl:cotRL}\textbf{CoT-RL EX Results in \%}}
\rowcolors{2}{gray!15}{white}
\resizebox{\columnwidth}{!}{%
\begin{tabular}{l|cc|cc|cc}
\hline 
Models & \multicolumn{2}{c|}{Spider(dev)} &  \multicolumn{2}{c|}{BIRD(dev)} & \multicolumn{2}{c}{Spider2.0--SQLite} \\ 
 & Gre & Maj & Gre & Maj & Gre & Maj \\
\hline
OmniSQL-7B\cite{li2025omnisql} & 81.2 & 81.6 &63.9  & 66.1 & 15.6$^*$  & 17.0$^*$\\
SQL-R1-7B\cite{ma2025sqlr1} & - & 87.6  & - & 66.6  & - &-\\
Arctic-Text2SQL-R1\cite{yao2025arcticsql} & - & - & 67.6  & 69.4  & 15.6  &-\\
\hline
PaVeRL-Spider & \textbf{83.4}  & 86.6  & N/A & N/A & N/A & N/A \\
PaVeRL-BIRD & N/A & N/A & 67.2  & 69.3  & N/A & N/A \\
PaVeRL-SynSQL10K & 78.2  & 81.7  & 60.6  & 64.7  & \textbf{17.0}  & \textbf{19.3}  \\
\hline
\end{tabular}}
\par\vspace{2pt}\footnotesize
$^*$Recalculated after database question update. Upper panel shows literature-reported accuracy. '-' indicates values not reported in literature. Spider-trained models tested only on Spider; BIRD-trained only on BIRD. N/A indicates values not applicable due to table structure.
\end{table}

% ===== Table 3: Mixed Dialects (keeps label tbl:mixRL) =====
\begin{table}[t]
\centering
\setlength{\tabcolsep}{3pt}
\caption{\label{tbl:mixRL}\textbf{PaVeRL with Mixed Dialects EX in \%}}
\rowcolors{2}{gray!15}{white}
\resizebox{\columnwidth}{!}{%
\begin{tabular}{l|c|ccc|cc}
\hline 
\multicolumn{2}{c|}{} & \multicolumn{3}{c|}{Spider2.0--SQLite} &  \multicolumn{2}{c}{in-house-MySQL}\\
\hline
\multicolumn{2}{c|}{Model} & EX & $\text{EX}_b$ & $\text{EX}_f$ & $\text{EX}_b$ & $\text{EX}_f$ \\
\hline
OmniSQL-7B \cite{li2025omnisql}  & Gre & 15.6 & 13.3 & 22.1  & 20.1 & 29.9 \\
\hline
PaVeRL-Mixed& Gre & 13.3 & 11.9 & 21.2  & 74.7 & 82.5  \\
            & Maj & 17.8 & 14.1 & 22.9  & 75.7 & 83.8  \\
\hline
\end{tabular}}
\par\vspace{2pt}\footnotesize
Note: our in-house data do not have public official evaluation metrics.
\end{table}

\subsection{CoT RL Pipeline}
\begin{figure*}[ht]
\centerline{\includegraphics[width=\textwidth]{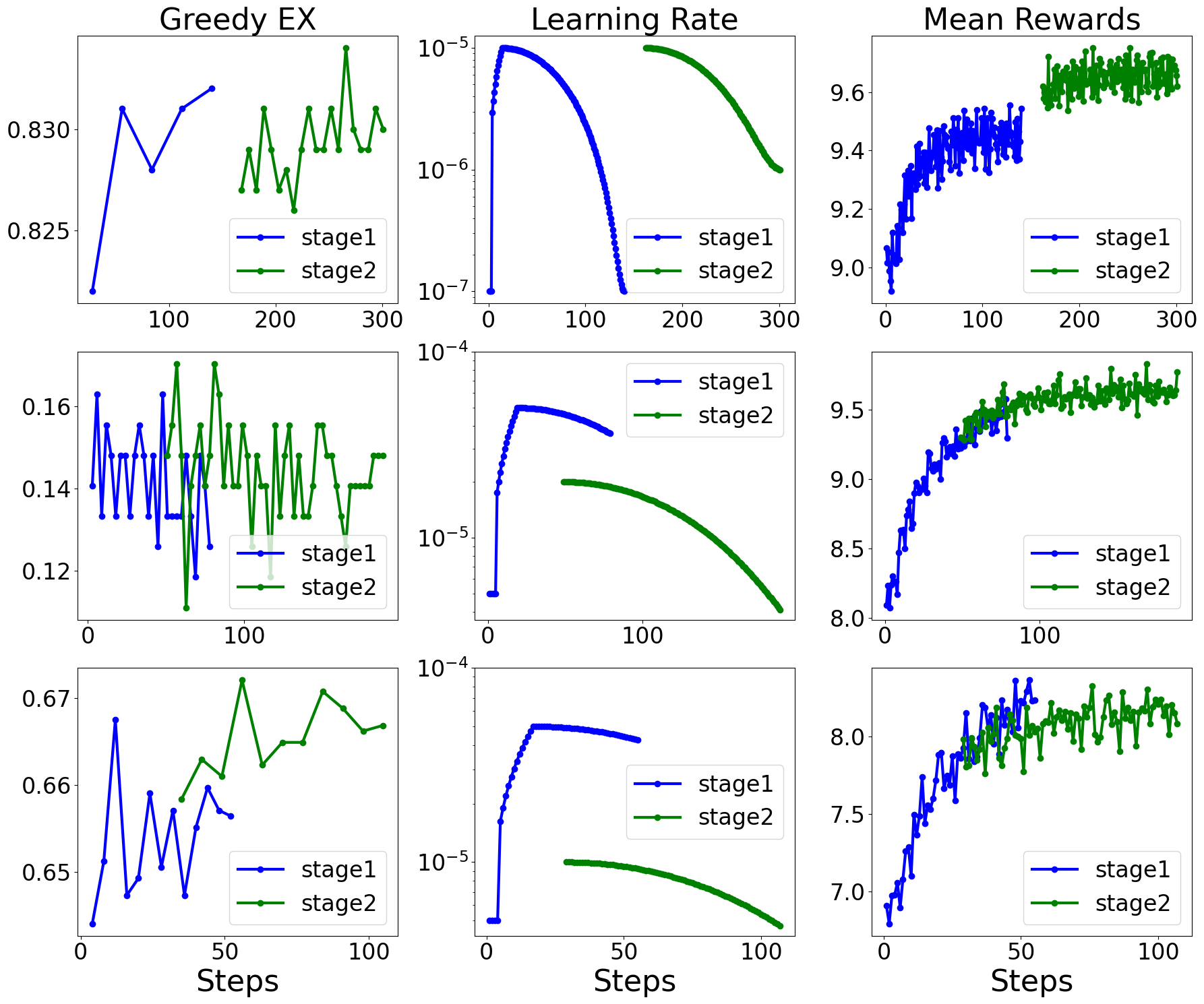}}
\caption{CoT RL training dynamics across two-stage GRPO optimization. Left: Greedy decoding execution accuracy (EX) progression showing model performance. Center: Learning rate schedules with stage 1 using warm up-linear-cosine decay, and stage 2 restarting from the best stage 1 checkpoint with adjusted rates based on convergence patterns. Right: Mean reward progression using the partial-match reward function, demonstrating improved signal density over binary rewards. Blue curves represent stage 1 training, green curves represent stage 2 training. Top to bottom: Spider\_dev, Spider2.0\textendash SQLite, and BIRD\_dev datasets. The two-stage approach enables convergence within 20 epochs while maintaining training stability.}
\label{fig:cotrl}
\end{figure*}

As described in Section~\ref{sec:cotrl}, we trained on Spider, BIRD, and SynSQL10K and evaluated on Spider, BIRD, and Spider2.0\textendash SQLite, respectively. Training was run on 8~H100 GPUs (96\,GB each). We use the VERL package~\cite{sheng2024hybridflow} with batch size 1024, temperature 0.8, and 10 rollouts. In Stage~1, the warm-up learning rate is $1\mathrm{e}{-7}$ or $5\mathrm{e}{-7}$; the maximum learning rate is $1\mathrm{e}{-5}$ or $5\mathrm{e}{-5}$ (depending on the run). Figure~\ref{fig:cotrl} shows greedy EX, the learning-rate schedule, and mean rewards across the two-stage GRPO training (blue: Stage~1; green: Stage~2).

Because SynSQL10K is general-purpose public data, we also evaluate the resulting model on Spider and BIRD. Execution accuracies of our best checkpoints versus recent RL Text-to-SQL systems (SQL-R1~\cite{ma2025sqlr1}, Arctic-Text2SQL-R1~\cite{yao2025arcticsql}) are summarized in Table~\ref{tbl:cotRL} (official EX only for brevity).

Overall, \emph{PaVeRL} CoT RL matches or exceeds SOTA RL systems, with clear gains on the more industrial Spider2.0\textendash SQLite benchmark.

\begin{figure*}[ht]
\centerline{\includegraphics[width=\textwidth]{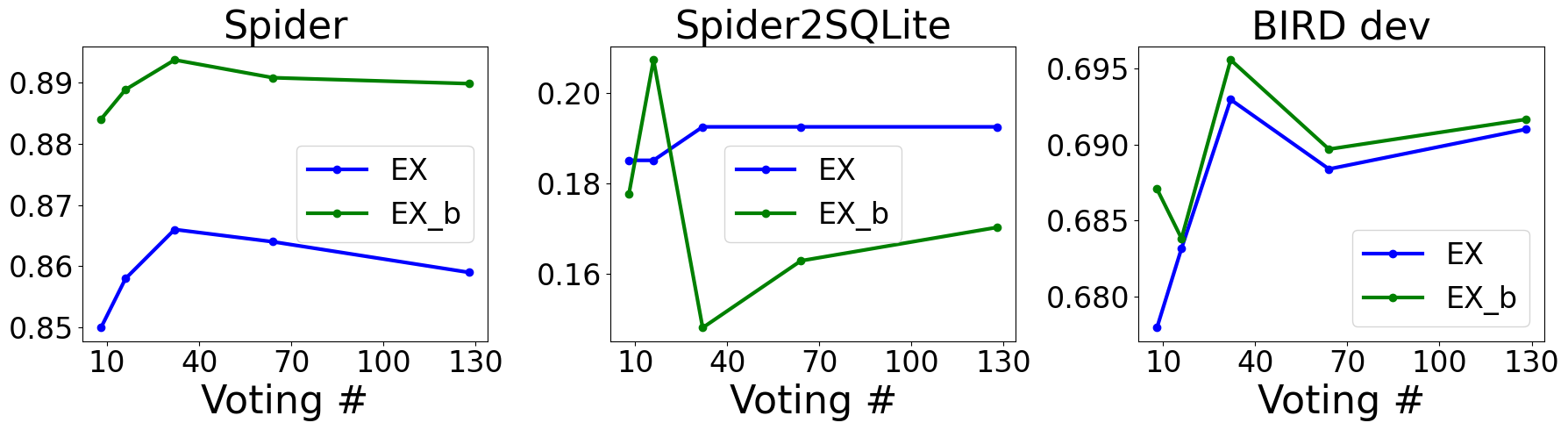}}
\caption{Impact of majority voting ensemble size on execution accuracy for CoT RL models. Results show official execution accuracy (EX, blue) and binary execution accuracy ($\text{EX}_b$, green) as a function of ensemble size (8, 16, 32, 64, 128 samples). Left to right: Spider\_dev, Spider2.0\textendash SQLite, and BIRD\_dev test sets. Note the different y-axis scales across datasets, reflecting varying baseline difficulty levels. The best-performing models for each dataset were selected based on highest majority voting EX scores.}
\label{fig:mv}
\end{figure*}

We additionally study how the number of samples used for majority voting affects accuracy. We test group sizes of 8, 16, 32, 64, and 128 and find that larger groups do not necessarily help; a size of \(\approx\)32 is typically optimal~(Figure~\ref{fig:mv}).

\subsection{Mixed Dialects CoT RL Training}
\label{sec:mix}
Training data for some SQL dialects are often limited. Although dialects differ in functions and types, the underlying reasoning is shared. We therefore test whether training on a data-rich dialect transfers to a low-resource dialect. Concretely, we randomly select 6{,}000 SynSQL10K SQLite samples and combine them with 2{,}000 proprietary MariaDB/MySQL samples (8{,}000 total) and run CoT RL as in Section~\ref{sec:cotrl}. Table~\ref{tbl:mixRL} compares results before and after RL on Spider2.0\textendash SQLite and an in-house MySQL test set. Performance on the rich SQLite test set does not degrade, while the low-resource MySQL set improves by roughly \(3\times\) in execution accuracy.

\section{Conclusion}
\label{sec:conclusion}
We presented \emph{PaVeRL\textendash SQL}, a practical Text\textendash to\textendash SQL framework that combines partial\textendash match rewards with two complementary tracks: a lightweight verbal self\textendash evaluation workflow and a compact CoT\textendash RL pipeline. By leveraging verifiable execution and denser feedback via $\text{EX}_b$ and $\text{EX}_f$, PaVeRL\textendash SQL improves learning stability and sample efficiency while remaining deployable under real\textendash world constraints. Across Spider, Spider2.0\textendash SQLite, and BIRD, we observe consistent accuracy gains, and mixed\textendash dialect training notably improves low\textendash resource dialects without degrading performance on richer ones.

Limitations include dependence on an executable environment and residual sensitivity of judge\textendash style scoring to backbone biases. Looking ahead, we will (i) enrich rewards beyond column\textendash level signals (e.g., tuple\textendash set comparisons, constraint satisfaction), (ii) strengthen cross\textendash model judging and calibration, and (iii) improve schema retrieval/compression for very large databases; we also plan to extend PaVeRL\textendash SQL to multi\textendash turn interactions and privacy\textendash constrained, on\textendash prem deployments.

%\section*{Acknowledgment}

\bibliographystyle{IEEEtran}
\bibliography{rlsqlref}

%\onecolumn
\appendices
%\clearpage
\section{Verbal RL Prompts} \label{app:vRLprompt}
The SQL generating prompt is the same as Prompt~\ref{sqlgenprompt}. We use the scoring prompt in a lightweight generate–judge loop that approximates GRPO without gradient updates: for each question we sample until we obtain \(K{=}10\) \emph{executable} candidates (deduplicated by normalized SQL), then query the same backbone to score the set and select the top candidate; ties are broken uniformly at random. The \texttt{<think>} block is ignored by our parser—only the numeric values inside \texttt{<scores>} are consumed—to limit susceptibility to prompt injection and formatting drift.

For reproducibility and stability, we clamp scores to \([0,1]\), optionally apply within-group z-score normalization before \(\arg\max\), and seed decoding. A light SQL linter and a dry-run executor filter non-executable outputs before judging, which reduces wasted judge capacity and aligns the loop with the executable-first reward used in training.

\begin{promptbox}{scoreprompt}{SQL Score Prompt}
Task Overview:

You are a scoring machine to make scores for \{len(all\_sqls)\} SQL scripts, based on understanding from input DATABASE SCHEMA, CONTEXT and QUESTION.
\\\\
Database Engine:\\
\{dialect\}
\\
Database Schema:\\
\{dbschema\}
\\
Context:\\
\{context\}
\\
Question:\\
\{question\}
\\\\
Instructions for scoring SQL scripts:\\
- Based on your understanding from input DATABASE SCHEMA, CONTEXT and QUESTION, please give score for each SQL script.\\
- Please compare all SQL scripts and give high score for SQL script that fully answers the input QUESTION.\\
- Please think through the steps with minimum number of tokens and the score should be between 0 and 1.\\
\\
\{sqls\}
\\
\\
Output Format:\\
In your answer, please enclose the thinking block with less than 1600 tokens followed by a code block with scores for each SQL script:
\begin{verbatim}
<think>
-- Your brief thinking
</think>
<scores>
{scores}
</scores>
\end{verbatim}

Please DO NOT generate any explanation to the final scores.
\end{promptbox}
 
\section{Database Schema String} \label{app:dbschema}
The following is an example database schema string that we use for both training and inference. Example values (MIN/MAX, top-3 modes) are computed by \emph{self-defined SQL functions} that scan each database to extract schema metadata and representative values. For long schemas, we always retain all PK/FK columns and any columns referenced by the gold SQL, and sample a small additional subset of columns to respect the context budget. Figure~\ref{dbschema} shows an example from the Spider dataset \texttt{school\_budget} database.

\begin{figure*}[ht]
\centering
\begingroup
% --- colors & spacing (tweak if desired) ---
\definecolor{shadecolor}{RGB}{244,248,255} % inner tint (soft blue)
\setlength{\FrameSep}{8pt}                 % padding
\setlength{\FrameRule}{0.5pt}              % border thickness

\begin{minipage}{0.86\textwidth} % <— controls width; 0.92\textwidth is centered
{\color{promptborder}% uses your existing border color
\begin{framed}
\color{black}
\begin{snugshade}
\begin{verbatim}
School 
CREATE TABLE School (
School_id  TEXT,  --Example  1, 2, 3,
School_name  TEXT,  --Example  Triton, New Prairie 1, LaVille,
Location  TEXT,  --Example  Walkerton, New Carlisle, Lakeville,
Mascot  TEXT,  --Example  Trojans, Redskins, Lions,
Enrollment  INT,  --MIN 287  MAX 852,
IHSAA_Class  TEXT,  --Example  AAA, AA, A,
IHSAA_Football_Class  TEXT,  --Example  AAA, A, AAAA,
County  TEXT,  --Example  50 Marshall, 71 St. Joseph, 75 Starke, 
PRIMARY KEY ("School_id")
);
budget 
CREATE TABLE budget (
School_id  INT,  --MIN 1  MAX 5,
Year  INT,  --MIN 1999  MAX 2006,
Budgeted  INT,  --MIN 3666  MAX 119527,
total_budget_percent_budgeted  REAL,  --MIN 1.3  MAX 2.4,
Invested  INT,  --MIN 2134  MAX 146102,
total_budget_percent_invested  REAL,  --MIN 2.0  MAX 2.7,
Budget_invested_percent  TEXT,  --Example  71.3, 42.9, 228.8, 
PRIMARY KEY("School_id","YEAR"),
FOREIGN KEY("School_id") REFERENCES "School"("School_id")
);
endowment 
CREATE TABLE endowment (
endowment_id  INT,  --MIN 1  MAX 11,
School_id  INT,  --MIN 1  MAX 8,
donator_name  TEXT,  --Example  Valverde, Santo Domingo Este, Santiago,
amount  REAL,  --MIN 8.33  MAX 9.83, 
PRIMARY KEY("endowment_id"),
FOREIGN KEY("School_id") REFERENCES "School"("School_id")
);
\end{verbatim}
\end{snugshade}
\end{framed}}
\end{minipage}
\endgroup

\caption{Example schema string for the Spider \texttt{school\_budget} database used in both training and inference. The string includes primary/foreign keys and lightweight profiles (MIN/MAX for numeric columns and top-3 modes for text) to guide joins and filters without leaking answers.}
\label{dbschema}
\end{figure*}

%\clearpage
%\section{Additional CoT RL EX results}
%\begin{figure*}[h]
%\centerline{\includegraphics[width=\textwidth]{output3.png}}
%\caption{Results of majority EX, greedy $\text{EX}_b$, and majority $\text{EX}_b$ (from left to right) as function of steps for 3 datasets, Spider\_dev, Spider2.0\textendash SQLite, and BIRD\_dev (from top to bottom). Blue and green dots are for stage 1 and 2, respectively.}
%\label{fig:exmv}
%\end{figure*}
\end{document}